\documentclass[letterpaper]{article}
\usepackage[draft]{aaai2027}
\usepackage[hyphens]{url}
\usepackage{graphicx}
\usepackage{natbib}
\usepackage{caption}
\usepackage{subcaption}
\usepackage{amsmath,amssymb,booktabs,multirow,xcolor}
\newtheorem{proposition}{Proposition}

\newtheorem{corollary}{Corollary}
\newtheorem{definition}{Definition}

\newcommand{\wx}[1]{\(\times\)#1}
\newcommand{\E}{\mathbb{E}}
\newcommand{\N}{\mathcal{N}}
\newcommand{\diag}{\operatorname{diag}}

\title{Width Can Connect Posterior Means Without Connecting Bayesian Predictions}
\author {
    Keumseo Ryum\textsuperscript{\rm 1},
    Joonhyuk Kang\textsuperscript{\rm 1}\corresponding
}
\affiliations {
    \textsuperscript{\rm 1}KAIST\\
    keumseo@kaist.ac.kr, jkang@kaist.ac.kr
    }

\begin{document}
\maketitle

\begin{abstract}
Independently trained deterministic networks can exhibit linear mode connectivity at large width even without permutation alignment. We ask whether width has the same effect in Bayesian neural networks by comparing the posterior mean network and the Bayesian model average (BMA) along the same Wasserstein paths between mean-field variational posteriors. For ResNet20 on CIFAR10, increasing width lowers the calibrated loss barrier of the posterior mean network, whereas the BMA barrier remains large. At widths \(\times16\) and \(\times32\), the posterior standard deviations change little along each path, while each hidden layer of the interpolated mean network produces a much smaller output relative to its input at the midpoint. This drop in layer gain reaches a median factor of about four and makes the same weight noise larger relative to the mean output by the same factor. We show that averaging the endpoint outputs reduces the gain similarly across widths, whereas the endpoint weight matrices retain less of their gain on midpoint inputs as width grows. We then propose a rescaling that shrinks the standard deviations between the endpoints by the predicted amount and prove that it keeps this relative noise at or below its endpoint value under idealized conditions. Experiments show that this rescaling removes most of the calibrated BMA barrier while preserving the mean path and both endpoint posteriors.
\end{abstract}

\begin{figure*}[t]
\centering
\begin{subfigure}[t]{0.32\textwidth}
\centering
\includegraphics[width=\linewidth]{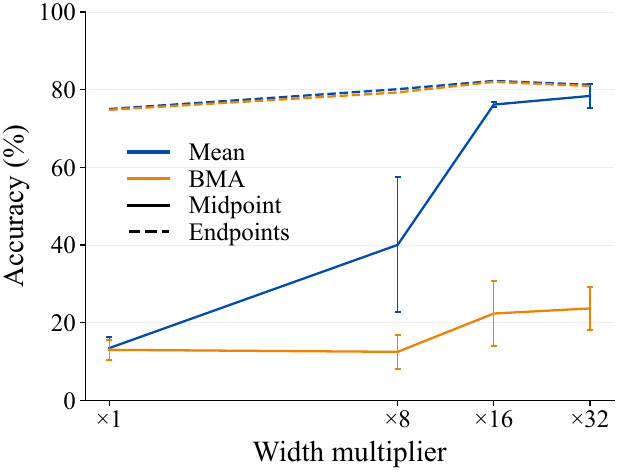}
\caption{Accuracy against width}
\label{fig:width:acc}
\end{subfigure}\hfill
\begin{subfigure}[t]{0.32\textwidth}
\centering
\includegraphics[width=\linewidth]{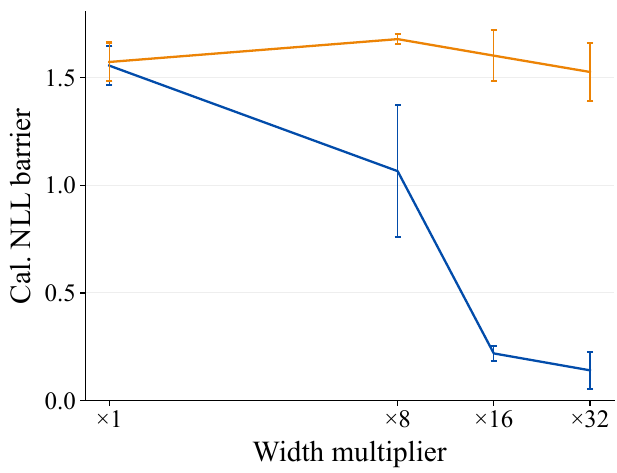}
\caption{Calibrated NLL barrier}
\label{fig:width:barrier}
\end{subfigure}\hfill
\begin{subfigure}[t]{0.32\textwidth}
\centering
\includegraphics[width=\linewidth]{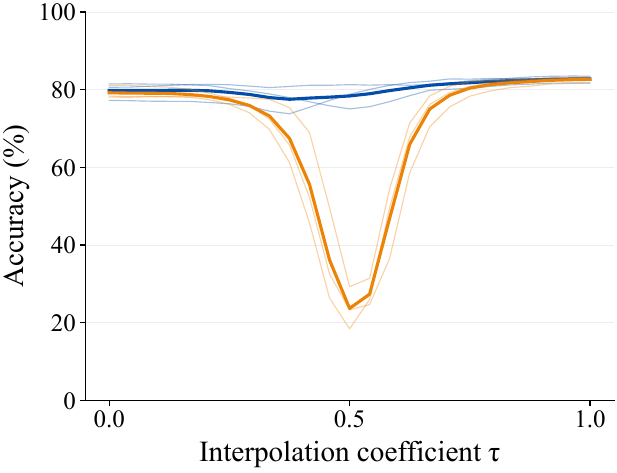}
\caption{Accuracy along the \wx{32} path}
\label{fig:width:path}
\end{subfigure}
\caption{Mean network and BMA on unaligned CIFAR10 ResNet20 paths. In (c), thin curves show individual pairs and the thick curve their mean. Error bars show the standard deviation over the three pairs.}
\label{fig:width}
\end{figure*}

\section{Introduction}

Linear mode connectivity (LMC) asks whether two trained networks can be joined by a straight path in parameter space with little increase in loss \citep{frankle2020linear}. LMC motivates weight averaging in model merging \citep{ainsworth2023git}, since averaging two linearly connected networks gives a single network that performs about as well as either endpoint. Independently trained networks are usually not linearly connected, partly because of the permutation symmetry of hidden units. Prior work on LMC found that aligning the units of the two networks removes much of the loss barrier \citep{entezari2022role,ainsworth2023git}. \citet{ito2026permutations} recently showed that increasing width can also induce connectivity without any alignment once the softmax temperature is calibrated. They explain this behavior with layerwise exponentially weighted connectivity (LEWC), under which each layer output of the interpolated network is a weighted sum of the corresponding outputs of the two endpoint networks.

Bayesian neural networks (BNNs) replace the single weight vector with a posterior distribution over the weights and predict with the Bayesian model average (BMA). The BMA can improve predictive performance over a single network \citep{izmailov2021bayesian}. Mean-field variational inference (MFVI) makes inference tractable by approximating the posterior with an independent Gaussian for every weight \citep{blundell2015weight}, so that each weight has a learned mean and standard deviation. For such posteriors, connectivity can be measured for the deterministic network at the posterior mean or for the BMA of networks sampled around that mean. The distinction matters when posteriors are combined by interpolating their parameters, since the predictions of a BNN come from the BMA.

\citet{rossi2023permutation} found that the BMA barrier between independently trained ResNet20 posteriors decreased with width only after permutation alignment. Their results leave open whether the unaligned mean networks remain poorly connected or whether width connects the means while a large barrier persists in the BMA. In this work, we distinguish these possibilities by evaluating both predictors along the same Wasserstein paths as width increases. We then trace the difference between the two predictors to the gain of individual layers and use this explanation to reduce the BMA barrier. Our contributions are as follows.

% \begin{itemize}
% \item We show that width can connect the posterior means of unaligned MFVI posteriors while the BMA on the same paths stays disconnected, so connectivity of the posterior means does not imply connectivity of the BMA (Section~\ref{sec:width}).
% \item We show that at large width the increase in relative sampling noise at the midpoint comes almost entirely from a loss of layer gain in the mean weights, and derive an exact decomposition of this loss that reduces to the LEWC case (Sections~\ref{sec:layer-noise} to~\ref{sec:measured-gain}).
% \item Using the measured midpoint noise ratio, we shrink the standard deviations inside the path and remove most of the calibrated BMA barrier while preserving both endpoint posteriors (Section~\ref{sec:interior-sd}).
% \end{itemize}

\begin{itemize}
\item We show that on unaligned Wasserstein paths between CIFAR10 ResNet20 posteriors, increasing width lowers the calibrated loss barrier of the posterior mean network while the BMA barrier remains large (Section~\ref{sec:width}).

\item We factor the midpoint drop in layer gain into a part from averaging the endpoint outputs and a part from applying the endpoint weight matrices to the midpoint input. Only the second part depends on width, and it rises to about two, its value under the conditions that \citet{ito2026permutations} use to explain width-induced connectivity (Sections~\ref{sec:layer-noise} to~\ref{sec:measured-gain}).

\item We propose a rescaling of the standard deviations that varies with the interpolation coefficient and prove that it keeps the layer noise ratio at or below its endpoint value along an idealized path. The rescaling leaves both endpoint posteriors unchanged and lowers the calibrated BMA barrier for every model pair at \wx{16} and \wx{32} (Section~\ref{sec:interior-sd}).
\end{itemize}

\section{Related Work}
\begin{figure}[t]
\centering
\begin{subfigure}[t]{\columnwidth}
\centering
\includegraphics[width=\linewidth]{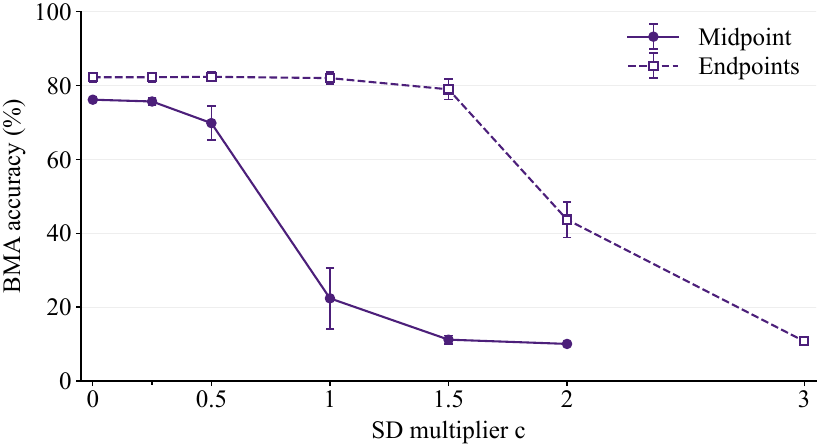}
\caption{Width \wx{16}}
\label{fig:scale:16}
\end{subfigure}

\begin{subfigure}[t]{\columnwidth}
\centering
\includegraphics[width=\linewidth]{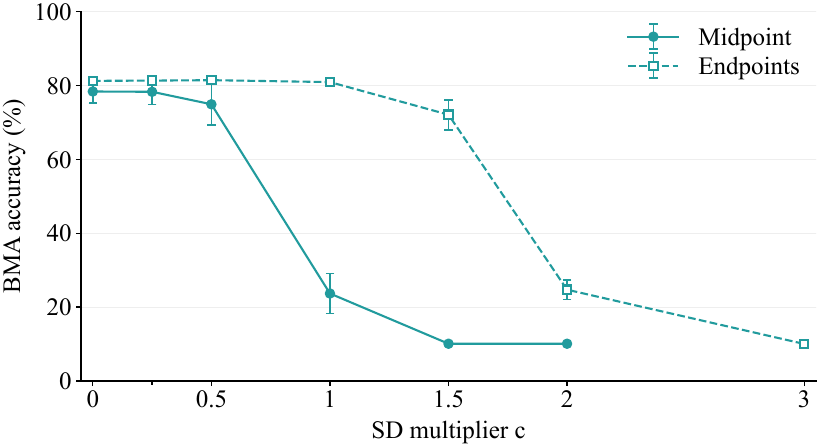}
\caption{Width \wx{32}}
\label{fig:scale:32}
\end{subfigure}
\caption{BMA accuracy against the standard deviation multiplier \(c\) at the midpoint and endpoints, where \(c=0\) gives the mean network and \(c=1\) the original standard deviations.}
\label{fig:scale}
\end{figure}
\paragraph{Mode connectivity}
\citet{garipov2018loss} and \citet{draxler2018essentially} find curved paths of low loss between independently trained networks, and \citet{frankle2020linear} measure connectivity through the barrier along straight paths. Permutation alignment reduces this barrier for sufficiently wide networks \citep{entezari2022role,ainsworth2023git}, and \citet{jordan2023repair} show that the hidden activations of interpolated networks lose variance and normalization restores it. \citet{zhou2023going} show that connectivity can also hold layer by layer in feature space. The drop in gain that we measure in the mean weights is related to the variance loss reported by \citet{jordan2023repair}, but it is defined per unit input energy and is unchanged when the layer input is multiplied by a constant.

\paragraph{Bayesian posteriors and symmetry}
\citet{izmailov2021bayesian} characterize BNN posteriors with Hamiltonian Monte Carlo on the full training set. \citet{gelberg2024variational} show that symmetries of the weight space cause failures of variational inference in Bayesian neural networks and propose permutation-invariant posteriors. \citet{rossi2023permutation} align diagonal Gaussian posteriors by permutation, report nearly zero BMA barriers after alignment, and conjecture that variational solutions become linearly connected once permutations are taken into account. They also report that naive interpolation collapses the variance of hidden activations and that alignment reduces this collapse. We study the same unaligned paths but evaluate the mean network and the BMA separately.

\paragraph{Mean-field variational inference}
MFVI \citep{blundell2015weight} restricts the posterior to independent Gaussians. Within this family, \citet{trippe2018overpruning} describe overpruning, in which many hidden units revert to the prior. For fully connected networks with odd activations and a homoscedastic Gaussian likelihood, \citet{coker2022wide} show that the optimal variational predictive distribution converges to the prior predictive distribution as width grows. \citet{foong2020expressiveness} show limits on the predictive uncertainty that mean-field posteriors can express. \citet{wenzel2020good} find that sharpening the posterior with a temperature below one during training improves the predictions of deep Bayesian networks. Unlike tempering during training, our scale intervention rescales the standard deviations of a trained posterior at evaluation time.

\section{Setup}

\subsection{Mean-field posteriors and interpolation}
Let \(f(x;\theta)\) denote the logits of a classifier with \(p(y\mid x,\theta)=\mathrm{softmax}(f(x;\theta))_y\). MFVI fits \(q(\theta)=\N(\mu,\diag(\sigma^2))\) by minimizing \(\E_{q}[-\log p(\mathcal{D}\mid\theta)]+\mathrm{KL}(q\,\|\,p)\) under a Gaussian prior \(p\). Given posteriors \(q_a\) and \(q_b\) trained from different seeds, we follow \citet{rossi2023permutation} and interpolate without permutation alignment along the Wasserstein \(W_2\) geodesic between diagonal Gaussians,
\begin{equation}
    \begin{aligned}
&q_\tau = \N\big(\mu_\tau,\diag(\sigma_\tau^2)\big), \\
&\mu_\tau = (1-\tau)\mu_a+\tau\mu_b, \\
&\sigma_\tau = (1-\tau)\sigma_a+\tau\sigma_b .
\end{aligned}
\end{equation}
For each \(\tau\), we evaluate the mean network \(p(y\mid x,\mu_\tau)\) and estimate the BMA by \(K^{-1}\sum_{k=1}^{K}p(y\mid x,\theta_k)\), with \(\theta_k\sim q_\tau\). We refer to \(\tau=1/2\) as the midpoint.

\subsection{Connectivity measures and calibration}
Following \citet{entezari2022role}, we measure the largest excess of the test loss \(\mathcal{L}(\tau)\) over the linear interpolation of the endpoint losses,
\begin{equation}
    \begin{aligned}
B=\max_{\tau\in[0,1]}\left\{\mathcal{L}(\tau)-\big[(1-\tau)\mathcal{L}(0)+\tau \mathcal{L}(1)\big]\right\},
    \end{aligned}
\end{equation}
where \(\mathcal{L}\) is the average test negative log likelihood (NLL) of either predictor. The calibrated barrier follows \citet{ito2026permutations} and applies temperature scaling \citep{guo2017calibration} to both predictors. For the mean network the temperature divides the logits \(f(x;\mu_\tau)\). For the BMA a common positive temperature divides the logits of every sample before the softmax probabilities are averaged,
\begin{equation}
    \begin{aligned}
\bar p_T(y\mid x)=\frac{1}{K}\sum_{k=1}^{K}\mathrm{softmax}\big(f(x;\theta_k)/T\big)_y .
    \end{aligned}
\end{equation}
Each temperature is fitted separately at every interpolation point by minimizing the predictive NLL on a calibration split of 2,000 test images. We report accuracy at temperature one and use the fitted temperatures only for the NLL and the barrier.

\section{Empirical Observations}
\label{sec:results}

\subsection{Experimental settings}
\label{sec:expsetup}
We train ResNet20 on CIFAR10 with width multipliers of \wx{1}, \wx{8}, \wx{16}, and \wx{32} using the filter response normalization (FRN) and thresholded linear units (TLUs) of \citet{singh2020filter}. Training follows the hyperparameters of \citet{rossi2023permutation} with 1000 epochs, no augmentation, and a Gaussian prior with variance 0.01. For each width, we train six posteriors from different seeds and interpolate within three disjoint pairs. We evaluate each path at 25 evenly spaced points with \(K=128\) posterior samples for the BMA, compute the barrier for each pair, and average it over the three pairs. We provide the training and evaluation details in Appendix~\ref{app:eval}.

\subsection{Mean and BMA barriers across widths}
\label{sec:width}
Figure~\ref{fig:width} compares the mean network and the BMA in midpoint accuracy and calibrated barrier across widths, and in accuracy along the \wx{32} paths. At \wx{1}, both predictors have large calibrated barriers and similarly low midpoint accuracy. As the width grows, the calibrated barrier of the mean network decreases as in the deterministic networks of \citet{ito2026permutations}, while the BMA barrier stays large. At \wx{32}, the BMA loses most of the midpoint accuracy of the mean network, whereas the two predictors have nearly the same endpoint accuracy. These results support the second possibility left open by \citet{rossi2023permutation}, in which width connects the posterior means while a large barrier persists in the BMA. Since the mean network is the BMA with every standard deviation set to zero, the excess of the BMA barrier over that of the mean network comes from sampling around the interpolated mean.

\subsection{Endpoint and midpoint sensitivity to posterior scale}
\label{sec:scale}
To compare the midpoint and endpoints over a range of sampling scales, we multiply all posterior standard deviations at the midpoint and at each endpoint by \(c\). Figure~\ref{fig:scale} shows that as \(c\) increases from zero to the trained scale, midpoint BMA accuracy falls sharply while endpoint accuracy remains close to that of the mean network at both widths.

\section{Analysis of the Predictive Barrier}
\label{sec:theory}
\subsection{Sampling noise and layer gain}
\label{sec:layer-noise}

\begin{figure}[t]
\centering
\includegraphics[width=0.6\columnwidth]{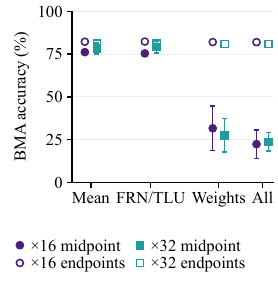}
\caption{BMA accuracy shift at \wx{16} and \wx{32} depending on which parameter group is sampled.}
\label{fig:types}
\end{figure}

Figure~\ref{fig:types} shows that at \wx{16} and \wx{32} sampling only the convolution and linear weights and biases accounts for most of the midpoint accuracy drop, whereas sampling only the FRN and TLU parameters barely changes midpoint accuracy. We therefore compare the perturbation from sampling the weights with the output of their means on the same input.

Consider a convolution or linear layer at either endpoint or at the midpoint, indexed by \(\tau\in\{a,b,1/2\}\), where \(a\) and \(b\) denote the endpoints at \(\tau=0\) and \(\tau=1\). Let \(h_\tau(x)\) be its input in the mean network, \(M_\tau\) its mean weight matrix, and \(\sigma_{\tau,ij}\) the posterior standard deviation of weight \(ij\). Writing the sampled matrix as \(W_\tau=M_\tau+\varepsilon_\tau\) splits the layer output \(W_\tau h_\tau\) into the mean output \(M_\tau h_\tau\) and the sampling noise \(\varepsilon_\tau h_\tau\). We measure both before the bias, normalization, and residual connection. Summing squared norms over evaluation inputs \(x\) and over spatial input patches for convolutions gives the input energy \(H_\tau\), the mean output energy \(Z_\tau\), and the expected sampling noise energy \(N_\tau\),
\begin{equation}
\begin{aligned}
    H_\tau &= \sum_x \|h_\tau(x)\|^2, \\
    Z_\tau &= \sum_x \|M_\tau h_\tau(x)\|^2, \\
    N_\tau &= \sum_x \mathbb{E}\|\varepsilon_\tau h_\tau(x)\|^2
            = \sum_{x,i,j}\sigma_{\tau,ij}^2 h_{\tau,j}(x)^2 ,
\end{aligned}
\label{eq:layer-energies}
\end{equation}
where the second expression for \(N_\tau\) holds because the entries of \(\varepsilon_\tau\) are independent with zero mean and variances \(\sigma_{\tau,ij}^2\). The ratio \(N_\tau/Z_\tau\) measures the sampling noise energy relative to the mean output energy. Because \(h_\tau\) is taken from the mean network, this ratio measures the noise injected at the layer and excludes noise propagated from earlier layers. We compare it at the midpoint with its value pooled over the two endpoints,
\begin{equation}
    \begin{aligned}
    D = \frac{N_{1/2}/Z_{1/2}}{(N_a+N_b)/(Z_a+Z_b)} .
    \end{aligned}
    \label{eq:noise-ratio}
\end{equation}
Dividing the output and noise energies by the input energy of the same layer factors \(D\) exactly as
\begin{equation}
    \begin{aligned}
    D &= Q\,\frac{N_{1/2}/H_{1/2}}{(N_a+N_b)/(H_a+H_b)}, \\
    Q &= \frac{(Z_a+Z_b)/(H_a+H_b)}{Z_{1/2}/H_{1/2}} .
    \end{aligned}
    \label{eq:dgain}
\end{equation}
The gain ratio \(Q\) compares the pooled endpoint gain with the midpoint gain, where the gain \(Z_\tau/H_\tau\) is the squared output norm of the mean weights per unit squared input norm. The remaining factor compares the sampling noise energy per unit input energy. If all weights in the layer share one standard deviation \(\sigma_0\) at both endpoints, the midpoint has the same standard deviation. For \(m\) output units this gives \(N_\tau/H_\tau=m\sigma_0^2\) at all three locations, so \(D=Q\). For the trained posteriors, \(D/Q\) gives the midpoint noise energy per unit input energy relative to its pooled endpoint value.

\subsection{Decomposition of the layer gain ratio}
\label{sec:gain-decomposition}

\begin{figure*}[t]
\centering
\includegraphics[width=0.9\textwidth]{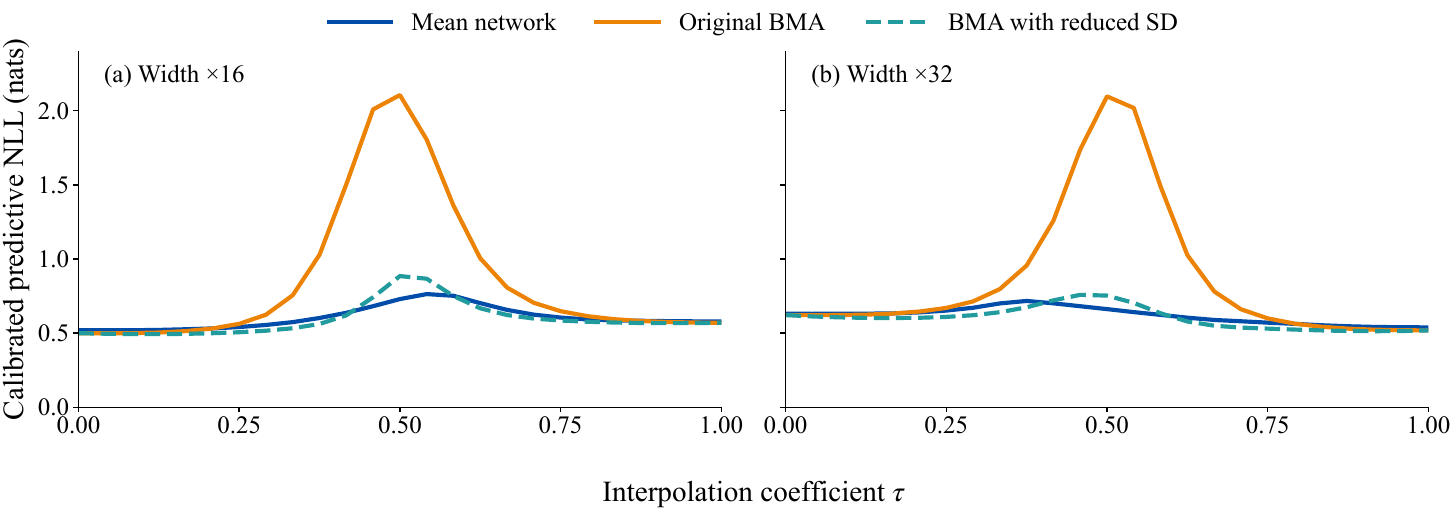}
\caption{Calibrated predictive NLL along the interpolation path for the mean network, the original BMA, and the BMA with standard deviations scaled by \(c(\tau)\) in \eqref{eq:sd-schedule}. Each curve averages three model pairs at 25 interpolation points.}
\label{fig:interp}
\end{figure*}

To decompose \(Q\), we first apply each endpoint matrix to the midpoint input and compare the resulting gains with the pooled endpoint gain. We then hold that input fixed and measure the further change in gain when the two outputs are averaged. Intuitively, this separates the gain ratio into two factors: $\Gamma$, which measures how the endpoint weights process the midpoint input relative to their own inputs, and $\eta$, which measures the alignment between the two resulting endpoint outputs.

\begin{proposition}
\label{prop:decomp}
Let \(Z_{1/2}^a=\sum_x\|M_a h_{1/2}(x)\|^2\) and
\(Z_{1/2}^b=\sum_x\|M_b h_{1/2}(x)\|^2\). Wherever the ratios are defined, the gain ratio satisfies
\begin{equation}
\begin{aligned}
    Q &= \frac{2\Gamma}{1+\eta},\\
    \Gamma &=
    \frac{(Z_a+Z_b)/(H_a+H_b)}
         {(Z_{1/2}^a+Z_{1/2}^b)/(2H_{1/2})},\\
    \eta &=
    \frac{2\sum_x\langle M_a h_{1/2}(x),M_b h_{1/2}(x)\rangle}
         {Z_{1/2}^a+Z_{1/2}^b}.
\end{aligned}
\label{eq:gain-decomp}
\end{equation}
\end{proposition}

\noindent\textit{Proof.}
Expanding the squared norm of
\(M_{1/2}h_{1/2}=(M_a h_{1/2}+M_b h_{1/2})/2\)
gives
\begin{equation}
\begin{aligned}
    Z_{1/2}=\tfrac14(Z_{1/2}^a+Z_{1/2}^b)(1+\eta).
\end{aligned}
\end{equation}
Dividing by \(H_{1/2}\) and substituting into \eqref{eq:dgain}
gives \(Q=2\Gamma/(1+\eta)\).\hfill\(\square\)

The factor \(\Gamma\) compares the pooled endpoint gain with the average gain of the two endpoint matrices on the midpoint input. The second factor \(2/(1+\eta)\) is the further drop in gain from averaging their outputs on that input and equals two when \(\eta=0\).

At the input convolution all three networks receive the same image \(x\), and \(h_{1/2}=h_a=h_b\). Proposition~\ref{prop:decomp} then gives \(\Gamma=1\) and \(Q=2/(1+\eta)\). This layer thus keeps only the averaging factor of the hidden layers. For each weight,
\begin{equation}
\begin{aligned}
    \left(\frac{\sigma_{a,ij}+\sigma_{b,ij}}{2}\right)^2
    \leq \frac{\sigma_{a,ij}^2+\sigma_{b,ij}^2}{2},
\end{aligned}
\end{equation}
and summing over the weights and input patches of the shared input gives \(N_{1/2}\leq(N_a+N_b)/2\) and hence \(D\leq Q\) at this layer.

The decomposition takes a closed form under the conditions that \citet{ito2026permutations} use to explain width-induced connectivity.

\begin{definition}[LEWC]
Let \(f_\ell(x;\theta)\) denote the output of layer \(\ell\). Two networks with parameters \(\theta_a\) and \(\theta_b\) satisfy LEWC at layer \(\ell\) if, for every input \(x\) and every \(\lambda\in[0,1]\),
\begin{equation}
\begin{aligned}
    f_\ell(x;(1-\lambda)\theta_a+\lambda\theta_b)
    ={}& (1-\lambda)^\ell f_\ell(x;\theta_a) \\
       &+ \lambda^\ell f_\ell(x;\theta_b).
\end{aligned}
\label{eq:lewc}
\end{equation}
\end{definition}

\citet{ito2026permutations} show that a ReLU network without biases satisfies LEWC at every layer when the interpolated preactivations satisfy a weak additivity condition and the weights of every hidden layer satisfy reciprocal orthogonality. Under reciprocal orthogonality, each endpoint weight matrix maps the input of the other network to zero. If the mean networks satisfy LEWC at layer \(\ell-1\), the midpoint input of layer \(\ell\) equals \(2^{-(\ell-1)}(h_a+h_b)\). Because \(\Gamma\) and \(\eta\) do not change when \(h_{1/2}\) is multiplied by a constant, the corollary below allows any nonzero scale \(\alpha\).

\begin{corollary}
\label{cor:lewc}
Define
\begin{equation}
\begin{aligned}
    \rho &=
    \frac{2\sum_x\langle h_a(x),h_b(x)\rangle}{H_a+H_b},\\
    \kappa &=
    \frac{2\sum_x\langle M_a h_a(x),M_b h_b(x)\rangle}{Z_a+Z_b}.
\end{aligned}
\label{eq:endpoint-products}
\end{equation}
Suppose that \(h_{1/2}(x)=\alpha\,(h_a(x)+h_b(x))\) for a scalar \(\alpha\neq0\) and that \(M_a h_b(x)=M_b h_a(x)=0\) for every \(x\). Then \(\Gamma=2(1+\rho)\) and \(\eta=\kappa\), so that, wherever the ratios are defined, \(Q=4(1+\rho)/(1+\kappa)\).
\end{corollary}

\noindent\textit{Proof.}
The assumptions give \(M_a h_{1/2}=\alpha M_a h_a\) and \(M_b h_{1/2}=\alpha M_b h_b\), so \(Z_{1/2}^a=\alpha^2 Z_a\), \(Z_{1/2}^b=\alpha^2 Z_b\), and the inner product in \(\eta\) equals \(\alpha^2\kappa(Z_a+Z_b)/2\). Expanding the squared norm of \(h_{1/2}\) gives \(H_{1/2}=\alpha^2(H_a+H_b)(1+\rho)\). Substituting into \eqref{eq:gain-decomp} gives \(\Gamma=2(1+\rho)\) and \(\eta=\kappa\).\hfill\(\square\)

Under these assumptions, each endpoint matrix responds only to the part of the midpoint input that comes from its own network. When \(\rho=\kappa=0\), the endpoint matrices therefore retain half their pooled gain on the midpoint input (\(\Gamma=2\)). Their outputs are orthogonal (\(\eta=0\)), so the gain of their average is one quarter of the pooled endpoint gain. Thus \(Q=4\), and \(D=4\) under a common weight standard deviation.

\subsection{Layer gain and midpoint accuracy}
\label{sec:measured-gain}

\begin{table}[t]
\centering
\small
\begin{tabular}{@{}lrrrr@{}}
\toprule
Width & \(\times1\) & \(\times8\) & \(\times16\) & \(\times32\) \\
\midrule
\(D\) & 2.75 & 3.72 & 3.94 & 3.81 \\
Max.\ \(\lvert D/Q-1\rvert\) (\%) & 8.1 & 1.7 & 0.8 & 0.6 \\
\midrule
\(\Gamma\) & 1.44 & 1.86 & 1.95 & 1.91 \\
\(\eta\) & 0.02 & 0.00 & 0.00 & 0.00 \\
\midrule
\(\rho\) & 0.18 & 0.05 & 0.04 & 0.04 \\
\(\lvert 4/Q-1\rvert\) (\%) & 46.5 & 11.5 & 6.2 & 6.1 \\
\(\lvert Q_{\mathrm{cor}}/Q-1\rvert\) (\%) & 69.5 & 18.1 & 8.9 & 9.8 \\
\bottomrule
\end{tabular}
\caption{Medians over 20 hidden layers and three model pairs at each width, except for the maximum in the second row. \(Q_{\mathrm{cor}}=4(1+\rho)/(1+\kappa)\) is the prediction of Corollary~\ref{cor:lewc} with the measured \(\rho\) and \(\kappa\).}
\label{tab:layer-statistics}
\end{table}

Table~\ref{tab:layer-statistics} shows that the median \(D\) is close to four at \wx{16} and \wx{32}. Because the trained standard deviations vary little within each hidden layer and between matched endpoint weights, \(D\) differs from \(Q\) by less than one percent in every hidden layer and pair at these widths. Most of the increase in noise relative to the mean output therefore comes from the lower gain of the interpolated weights.

Since the median \(\eta\) stays near zero, averaging the endpoint outputs contributes a factor close to two at every width and \(\Gamma\) carries the width dependence. At \wx{16} and \wx{32} the measured factors are close to \(\Gamma=2\) and \(\eta=0\), the values that Corollary~\ref{cor:lewc} gives when \(\rho=\kappa=0\). Four predicts \(Q\) more accurately than the corollary with the measured \(\rho\) and \(\kappa\) at every width. The measured \(\rho\) stays positive at large width and raises the corollary prediction above four, whereas the measured \(Q\) stays slightly below four. While the trained networks only approximate the strict assumptions of Corollary ~\ref{cor:lewc}, it serves as a robust idealized reference for understanding the underlying gain dynamics.

Doubling the endpoint standard deviations multiplies \(N/Z\) by four, close to the median \(D\) at \wx{16} and \wx{32}. Table~\ref{tab:scale} shows that matching the layer noise ratio in this way makes the two drops similar at \wx{32}, whereas the midpoint still loses substantially more accuracy at \wx{16} despite a similar median \(D\). This discrepancy at \wx{16} suggests that when the model capacity is smaller, the network is less robust to the accumulated noise propagated through the modified layer gains, leading to a steeper degradation in the final BMA accuracy.

\subsection{Reducing interior standard deviations}
\label{sec:interior-sd}
Scaling the midpoint standard deviations by \(0.5\) divides \(D\) by four, bringing its median close to one at \wx{16} and \wx{32}. Figure~\ref{fig:scale} shows that this scaling largely closes the midpoint accuracy gap at \wx{32}. To test whether this reduction also lowers the BMA barrier along the whole path, we keep the interpolated means fixed and multiply all posterior standard deviations by
\begin{equation}
    \begin{aligned}
    c(\tau)=1-2\tau(1-\tau)=(1-\tau)^2+\tau^2.
    \end{aligned}
    \label{eq:sd-schedule}
\end{equation}
This is the unique quadratic that equals one at both endpoints and one half at the midpoint. The adjusted path therefore connects the same two posteriors, although it is no longer the Wasserstein geodesic. Let \(D(\tau)\) denote \eqref{eq:noise-ratio} with the midpoint replaced by \(\tau\). If the mean networks satisfy LEWC at layer \(\ell-1\), the input of layer \(\ell\) takes the form in the proposition below with \(n=\ell-1\) and \(\alpha_\tau=1\). The following result extends the midpoint argument to the whole path under idealized conditions.

\begin{proposition}
\label{prop:schedule}
Suppose that, at a hidden layer with \(m\) output units and for every \(\tau\in[0,1]\),
\begin{equation}
\label{eq:path-input}
\begin{aligned}
    h_\tau(x)=\alpha_\tau\big((1-\tau)^n h_a(x)+\tau^n h_b(x)\big)
\end{aligned}
\end{equation}
with an integer \(n\geq1\) and \(\alpha_\tau\neq0\). Suppose also that (i) \(M_a h_b(x)=M_b h_a(x)=0\) for every \(x\), (ii) \(\rho=\kappa=0\) with \(H_a=H_b\) and \(Z_a=Z_b\), and (iii) every weight at both endpoints has standard deviation \(\sigma_0\). If the standard deviations at \(\tau\) are multiplied by \(c(\tau)\) in \eqref{eq:sd-schedule}, then \(D(\tau)\leq1\) on \([0,1]\), with equality only at \(\tau\in\{0,1/2,1\}\).
\end{proposition}

\noindent\textit{Proof.}
Since \(M_\tau=(1-\tau)M_a+\tau M_b\), assumption (i) gives
\begin{equation}
\begin{aligned}
    M_\tau h_\tau=\alpha_\tau\big((1-\tau)^{n+1}M_a h_a+\tau^{n+1}M_b h_b\big).
\end{aligned}
\end{equation}
Assumption (iii) gives \(N_\tau/H_\tau=m\,c(\tau)^2\sigma_0^2\). Expanding \(Z_\tau\) and \(H_\tau\) under (ii) then gives
\begin{equation}
\begin{aligned}
    D(\tau) &= \frac{c(\tau)^2}{L_{n+1}(\tau)}, \\
    L_r(\tau) &= \frac{(1-\tau)^{2r}+\tau^{2r}}{(1-\tau)^{2r-2}+\tau^{2r-2}} .
\end{aligned}
\end{equation}
With \(u=(1-\tau)^2\) and \(v=\tau^2\), every integer \(r\geq2\) satisfies
\begin{equation}
\begin{aligned}
    L_{r+1}(\tau)-L_r(\tau)=\frac{(uv)^{r-1}(u-v)^2}{(u^r+v^r)(u^{r-1}+v^{r-1})}\geq0,
\end{aligned}
\end{equation}
and summing these differences gives \(L_{n+1}\geq L_2\). With \(s=\tau(1-\tau)\in[0,1/4]\) and \(c(\tau)=1-2s\),
\begin{equation}
\begin{aligned}
    L_2(\tau)-c(\tau)^2=\frac{2s(1-s)(1-4s)}{1-2s}\geq0,
\end{aligned}
\end{equation}
with equality only at \(s\in\{0,1/4\}\). Together with \(L_{n+1}\geq L_2\), this gives \(D(\tau)\leq1\). At \(s\in\{0,1/4\}\), either \(uv=0\) or \(u=v\). Then \(L_{n+1}=L_2=c^2\) and \(D=1\).\hfill\(\square\)

Because the bound holds for every integer \(n\geq1\), the same rescaling applies to every hidden layer. At the midpoint \(D(1/2)=1\) for every \(n\), and the rescaling therefore shrinks the standard deviations there no more than needed.

Figure~\ref{fig:interp} shows that with the original standard deviations, calibrated BMA loss peaks near the middle of the path at both widths. Reducing the interior standard deviations brings the BMA curve closer to the mean network curve and lowers the barrier for all three pairs at each width. This rescaling removes most of the average calibrated BMA barrier, but the remaining barrier stays above that of the mean network.

\begin{table}[t]
\centering
\small
\begin{tabular}{@{}lcc@{}}
\toprule
Width & \(c'=1\) & \(c'=2\) \\
\midrule
\wx{16} & \(53.5\pm8.0\) & \(15.2\pm9.1\) \\
\wx{32} & \(54.4\pm6.2\) & \(-1.9\pm4.1\) \\
\bottomrule
\end{tabular}
\caption{Midpoint accuracy drop at \(c=1\) minus endpoint accuracy drop at \(c'\), in percentage points. Each drop is measured from the corresponding mean network. Values are the mean and standard deviation over three model pairs.}
\label{tab:scale}
\end{table}

\begin{figure}[t]
\centering
\begin{subfigure}[t]{0.49\columnwidth}
\centering
\includegraphics[width=\linewidth]{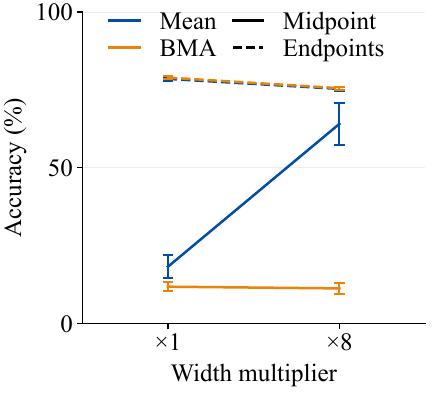}
\caption{Accuracy}
\label{fig:other:bn}
\end{subfigure}\hfill
\begin{subfigure}[t]{0.49\columnwidth}
\centering
\includegraphics[width=\linewidth]{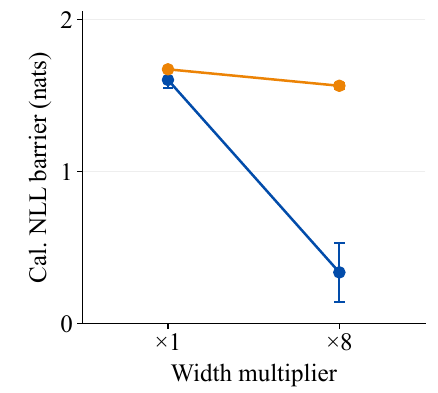}
\caption{Calibrated NLL barrier}
\label{fig:other:bnbar}
\end{subfigure}
\caption{Mean network and BMA for BN ResNet20 on unaligned CIFAR10 paths.}
\label{fig:other}
\end{figure}

\section{Discussion}
\label{sec:additional}
\paragraph{Other settings}
We also train BatchNorm (BN) ResNet20 models on CIFAR10 for 120 epochs with Adam and data augmentation. Figure~\ref{fig:other} shows accuracy and calibrated barrier against width for these models. From \wx{1} to \wx{8}, the mean network at the midpoint gains accuracy and its calibrated barrier drops sharply, while the BMA stays near chance accuracy with a barrier similar to its value at \wx{1}. Because the BN statistics are recomputed for each sampled network, this result shows that the separation persists even when normalization adapts to each sample.

\paragraph{Relation to alignment}
The decomposition also suggests why permutation alignment reduces the BMA barrier. When alignment brings the endpoints close, each endpoint matrix acts on the midpoint input much as on its own input and the two outputs nearly coincide. Both \(\Gamma\) and \(\eta\) then approach one, and \(Q=2\Gamma/(1+\eta)\) approaches one as well. This agrees with the small BMA barriers that \citet{rossi2023permutation} report after alignment.

\paragraph{Limitations}
We evaluate the layerwise gain decomposition on three pairs of FRN ResNet20 models per width on CIFAR10. Under the training procedure of \citet{rossi2023permutation}, the trained standard deviations vary little within each layer at \wx{16} and \wx{32}, and in two control runs at \wx{16} their median over hidden layers is nearly identical to the value obtained using only the KL term. For posteriors with more heterogeneous weight scales, both the size of the BMA barrier and the agreement between \(D\) and \(Q\) remain open questions.

\section{Conclusion}
For wide FRN ResNet20 posteriors on CIFAR10, a low calibrated barrier between the posterior means does not imply a low barrier for the BMA along the same path. At the midpoint, the interpolated mean weights have a much lower layer gain than at the endpoints, while the sampling noise per unit input energy stays nearly unchanged. Averaging the endpoint outputs reduces the gain at every width, and the further drop from applying the endpoint weights to the midpoint input grows as the mean networks become connected. The BN results show that recomputing the normalization statistics for each sampled network does not remove this separation. The proposed rescaling of the standard deviations bounds the layer noise ratio under idealized conditions and leaves a much smaller calibrated BMA barrier. The gap that remains in the matched noise comparison at \wx{16} indicates that layerwise noise ratios do not fully determine the BMA loss, likely due to the varying sensitivity of different network capacities to propagated errors. Future work will examine whether this separation persists under other approximate inference methods that learn more heterogeneous standard deviations. The layer gain of the interpolated means may also guide how the standard deviations are set when Bayesian models are merged by interpolating their parameters.

\bibliography{aaai2027}
% \bibliographystyle{aaai2027}  % uncomment only if the style file does not set it

%% =====================================================================
\appendix

\section{Training and Evaluation Details}
\label{app:eval}

\paragraph{FRN training and posterior scale}
We train the FRN networks with batches of 500 using SGD with momentum 0.9 and a constant learning rate of \(10^{-6}\). The objective is \(N\widehat{\mathrm{CE}}+\mathrm{KL}\), where \(\widehat{\mathrm{CE}}\) is the minibatch average cross-entropy, with \(N=50{,}000\) training images and one sampled network per step. We parameterize the posterior standard deviations with a softplus and initialize them at 0.01. After pooling the six endpoint checkpoints at each width, the median standard deviation in every hidden layer rounds to 0.0259 at both \(\times16\) and \(\times32\). In two control runs at \(\times16\), the median over hidden layers is within 0.1\% of the value obtained by applying the same SGD updates using only the KL term.

\paragraph{BN training}
We train the BN models for 120 epochs with Adam, a learning rate of 0.001, and batches of 500. We use a Gaussian prior with variance 0.01, initialize the posterior standard deviations at 0.01, and augment the training images with horizontal flips and random crops with padding 4. We treat the BN affine parameters as point estimates and recompute the BN statistics for each sampled network on all 50,000 training images.

\paragraph{Calibration and posterior sampling}
We use 2,000 test images to fit temperatures and the remaining 8,000 images for evaluation. We fit \(T\in[10^{-3},10^3]\) separately for each predictor, interpolation point, and scale condition. Reported endpoint values average the two endpoints of each pair. Increasing the number of posterior samples from 128 to 512 changes midpoint accuracy by at most 3.33 percentage points at \wx{16} and \wx{32}. The barriers of the mean network and the BMA also remain separated without temperature calibration, and rescaling by \(c(\tau)\) in \eqref{eq:sd-schedule} lowers the uncalibrated BMA barrier from 1.660 to 0.904 nats at \wx{16} and from 1.594 to 0.723 nats at \wx{32}.

\paragraph{Layer measurements}
We compute \(H_\tau\), \(Z_\tau\), and \(N_\tau\) on the first 1,000 evaluation images using one forward pass of the mean network at each endpoint and at the midpoint. For convolutions, we unfold layer inputs into spatial patches. We measure 20 hidden layers per model pair, including the two projection shortcuts and excluding the input convolution and classifier. At \(\times16\) and \(\times32\), the median relative difference between matched endpoint standard deviations is at most 0.079\% in any hidden layer, and the coefficient of variation within a layer is at most 0.85\%.

\end{document}